\documentclass[runningheads]{llncs}
\usepackage{graphicx}
\usepackage{subfigure}

\usepackage{diagbox}
\usepackage[misc]{ifsym}
\usepackage{float}

\begin{document}
\title{A Causal Disentangled Multi-Granularity Graph Classification Method}

\author{{Yuan Li\inst{1,2}}
\and
%
Li Liu\inst{1,2}  \and Penggang Chen\inst{1,2} \and Youmin Zhang\inst{1,2} \and Guoyin Wang\inst{1,2}$^{\textrm{(\Letter)}}$ }
\authorrunning{Y. Li et al.}

\institute{Chongqing Key Laboratory of Computational Intelligence,\and
Key Laboratory of Cyberspace Big Data Intelligent Security, Ministry of Education,\\
Chongqing University of Posts and Telecommunications,\\
Chongqing 400065, the People’s Republic of China\\
\email{D190201011@stu.cqupt.edu.cn},
\email{liliu@cqupt.edu.cn},
\email{hanoicpg@163.com},
\email{ymzhang0103@hotmail.com},
\email{wanggy@cqupt.edu.cn}}

\maketitle 

\begin{abstract}
Graph data widely exists in real life, with large amounts of data and complex structures. It is necessary to map graph data to low-dimensional embedding. Graph classification, a critical graph task, mainly relies on identifying the important substructures within the graph. At present, some graph classification methods do not combine the multi-granularity characteristics of graph data. This lack of granularity distinction in modeling leads to a conflation of key information and false correlations within the model. So, achieving the desired goal of a credible and interpretable model becomes challenging. This paper proposes a causal disentangled multi-granularity graph representation learning method (CDM-GNN) to solve this challenge. The CDM-GNN model disentangles the important substructures and bias parts within the graph from a multi-granularity perspective. The disentanglement of the CDM-GNN model reveals important and bias parts, forming the foundation for its classification task, specifically, model interpretations. The CDM-GNN model exhibits strong classification performance and generates explanatory outcomes aligning with human cognitive patterns. In order to verify the effectiveness of the model, this paper compares the three real-world datasets MUTAG, PTC, and IMDM-M. Six state-of-the-art models, namely GCN, GAT, Top-k, ASAPool, SUGAR, and SAT are employed for comparison purposes. Additionally, a qualitative analysis of the interpretation results is conducted.

\end{abstract}

\section{Introduction}
Graph data is characterized by complex structures and vast amounts of data, widely prevalent in our daily lives. Therefore, it is crucial to map graph data into low-dimensional embedding. 
Among various graph downstream tasks, graph classification is an essential task. Examples of such tasks include superpixel graph classification \cite{DBLP:conf/iclr/HendrycksD19}, molecular graph property prediction \cite{DBLP:conf/nips/HuFZDRLCL20}, and more.

In the research on graph classification, representation learning is an important approach for data analysis. 
Graph data, inherently possessing multiple levels of granularity, comprises nodes at a fine-grained level, coarser-grained substructures. For example, in the Tox21 dataset, atoms are fine-grained, functional groups are coarse-grained. This dataset compounds containing the azo functional group that are associated with carcinogenic and mutagenic properties \cite{fang2023knowledge}.
In graph classification, however, traditional representation learning methods overlook the multi-granularity of graph data, and these methods' interpretability is limited.
In graph classification, the outcome is primarily determined by certain important substructures \cite{DBLP:conf/aistats/LucicHTRS22}. Indeed, current graph classification methods do not consider the inherent multi-granularity nature of graph data. This lack of granularity differentiation during modeling leads to the mixing of critical information and false correlations within the models. As a result, it becomes challenging to accurately distinguish and achieve the goal of building interpretable models. 

In summary, there is a need to construct a substructure recognition model that takes into account multi-granularity graph data for modeling the graph representations. Therefore, this paper attempts to build a causal disentangled GNN model based on the idea of multi-granularity \cite{wang2017dgcc}. This model can disentangle the important substructures and bias parts in the graph from the multi-granularity perspective, and then conduct representation learning and classification for the entire graph.

Specifically, this paper designs a causal disentangled multi-granularity graph representation learning method (CDM-GNN). 
First, from a fine-grained perspective, this paper uses the feature and topological information of nodes to build a mask describing the closeness between nodes.
Next, from a coarser-grained perspective, CDM-GNN uses this mask to disentangle the important substructures and bias parts. It obtains the reason for learning the current representation, which is the interpretable result.
Subsequently, the masked graph is input into each slice layer to disentangle the key substructures from the false association relationship. 
Increase the depth of the model, gradually transitioning from fine-grained to coarse-grained, and expanding global information.
Finally, this paper learns an adaptive weight for each layer of slice results and adaptively fuses the results of each layer. Then it obtains the final representation for graph classification.

The main contributions of this work are summarized as follows:

1. The CDM-GNN considers the multi-granularity characteristics of graph data. It models through granularity transformation, fully taking into account the information at different granularities and their fusion.

2. The proposed model is capable of disentangling the key substructures and bais parts associative relationships in the graph while providing corresponding explanations.

3. Compared with six state-of-the-art models, namely GCN, GAT, Top-k, ASAPool, SUGAR, and SAT in MUTAG, PTC, and IMDB-M, the CDM-GNN model achieves better graph classification results.

\section{Related Work}
\subsection{Graph classification representation learning}
To obtain continuous low-dimensional embedding, graph representation learning aims to map non-Euclidean data into a low-dimensional representation space. For graph classification, there are two categories of research methods. One category is similarity-based graph classification methods, including graph kernel methods and graph matching methods. However, these methods are often inflexible and computationally expensive. In these methods, the process of graph feature extraction and graph classification is independent, which limits optimization for specific tasks.
The other category is based on GNNs. When applied to graph classification problems, GCN \cite{DBLP:conf/iclr/KipfW17} and GAT \cite{DBLP:conf/iclr/VelickovicCCRLB18} perform graph classification through convolution and pooling operations. Pooling is the process of graph coarsening, where the operation progressively aggregates fine-grained nodes. Subsequent research has also introduced changes to the pooling operation. For example, SAT \cite{DBLP:conf/icml/ChenOB22} proposes a graph transformer method used in pooling. 
\subsection{Disentangled learning}

The idea of disentangling initially originated from Bengio et al. \cite{DBLP:journals/pami/BengioCV13} and is primarily focused on computer vision \cite{DBLP:conf/nips/HsiehLHLN18}. However, some researchers have extended this concept to graphs. DisenGCN model \cite{DBLP:conf/icml/Ma0KW019} introduces a neighborhood routing mechanism to disentangle the various latent factors behind interactions in the graph.  
The IPGDN model \cite{DBLP:conf/aaai/LiuWWX20}, based on DisenGCN, add the Hilbert Schmidt Independence Criterion (HSIC) to further enhance the independence between different modules.
Based on the routing mechanism, the authors demonstrate the user-item relationship at the granularity of user intent and disentangle these intents in the representations of users and items \cite{DBLP:conf/sigir/WangJZ0XC20}. However, this method mainly focuses on bipartite graphs and may not be suitable for more complex graph structures. It lacks scalability.
In the context of knowledge graphs with richer types of relationships, some works consider leveraging relationship information in the process of disentangled representation learning. For example, they guide the disentangled representation of entity nodes based on the semantics of relationships \cite{zhang2021knowledge}. However, these methods overlook the information from different types of relationship edges.
These studies have successfully achieved disentangled. But, their primary emphasis is on manipulating the intermediate hidden layer states, which poses challenges in comprehending the structure of the graph. Consequently, their interpretability from a human perspective is limited.
\subsection{Interpretability method}

Despite the excellent representation capability of GNN models, their learning process is often opaque and difficult for humans to understand. To address this issue, some researchers have proposed post-hoc methods for explaining. GNNExplainer model \cite{DBLP:conf/nips/YingBYZL19} learns masks for the adjacency matrix and node features to identify important substructures. The PGExplainer model \cite{DBLP:conf/nips/LuoCXYZC020} attempts to learn an MLP function to mask edges in the graph, incorporating sparsity and continuity constraints in the model to obtain the final explanations. The post-hoc methods for explaining can only understand the model, not adjust the model. 
Another category of GNN explanation methods involves constructing self-explainable GNNs. Compared with post-hoc methods, this kind of model not only provides predictions but also offers explanations for the reasons behind those predictions. It can guide the model to some extent. Some self-explainable models require prior knowledge. For example, KerGNNs \cite{DBLP:conf/aaai/FengY0T22} is a subgraph-based node aggregation algorithm that manually constructs graph kernel functions to compare the similarity between graph filters and input subgraphs. The trained graph filters are also visualized and used as the model's explanation, which is then integrated into the GNN.
In addition to explaining isomorphic GNN models, researchers have also explored self-explainable models for heterogeneous GNN models, such as Knowledge Router \cite{DBLP:conf/iclr/CucalaGKM22}. However, these self-explainable GNNs have not extensively considered the issue of the multi-granularity structure of graphs.

\section{Preliminaries}
\subsection{Notations}
Let $G=\left ( V, E \right )$ be a graph, where $V$ is the node set, and $E$ is the edge set. The $A\in \left \{ 0,1 \right \} ^{\left | N \right |\times \left | N \right |  } $ is defined as the adjacency matrix of graph $G$. If there is an edge between node $i$ and node $j$, the $A_{ij}=1$; otherwise,$A_{ij}=0$. The $X$ is defined as the features matrix. $X\in R^{\left | V \right | \times F} $ represents the features of each node, where $F$ denotes the feature dimension for each node. The neighbour of node $i$ is $N_i$. The true label set is $Y$. 
\subsection{A Causal View on GNNs}
We analyze this problem using the Structural Causal Model (SCM). Figure 1 illustrates the five components. Z: input graph data, Y: labels, B: bias part in the graph, C: important substructures, and E: learned embedding by GNN model.

\begin{figure}[h]
    \centering
    \includegraphics[width=\textwidth]{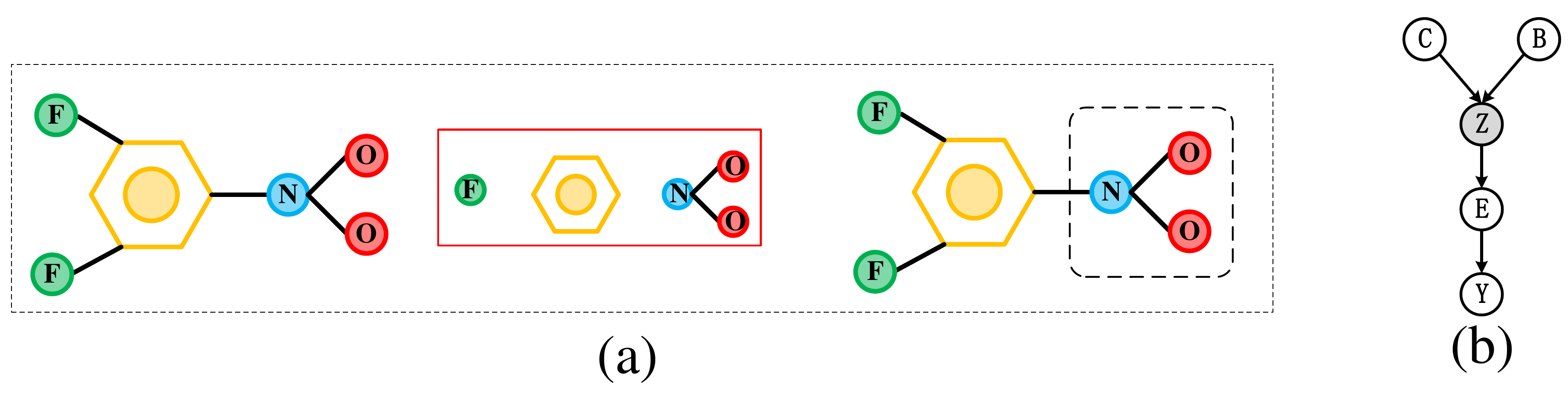}
    \caption{(a) A real example of MUTAG dataset. A molecular diagram consists of a carbon ring, $F$ atomic, and $NO_{2}$. The $NO_{2}$ atomic group determines to have mutagenicity, while others (bias part) do not determine this property. (b) Causal view of graph classification. The $NO_{2}$ atomic group is the C, and others are the bias parts B.}
    \label{Figure1}
\end{figure}

$C\to Z\gets B$: Z is composed of B and C.

$C\to Z\to E \to Y$: The structure of C is learned through GNNs and represented as E. Then, establish a causal correlation between C and Y.

$B\to Z\to E \to Y$: Due to the confusion between B and C, it affects the representation E obtained from GNNs, which also impacts the prediction of Y. Consequently, a spurious correlation is formed, leading to misleading predictions.

\section{Proposed Method}
This paper employs a multi-granularity approach for modeling. The CDM-GNN is introduced with an overall framework illustrated in Figure 2.
\begin{figure}
    \centering
    \includegraphics[width=\textwidth]{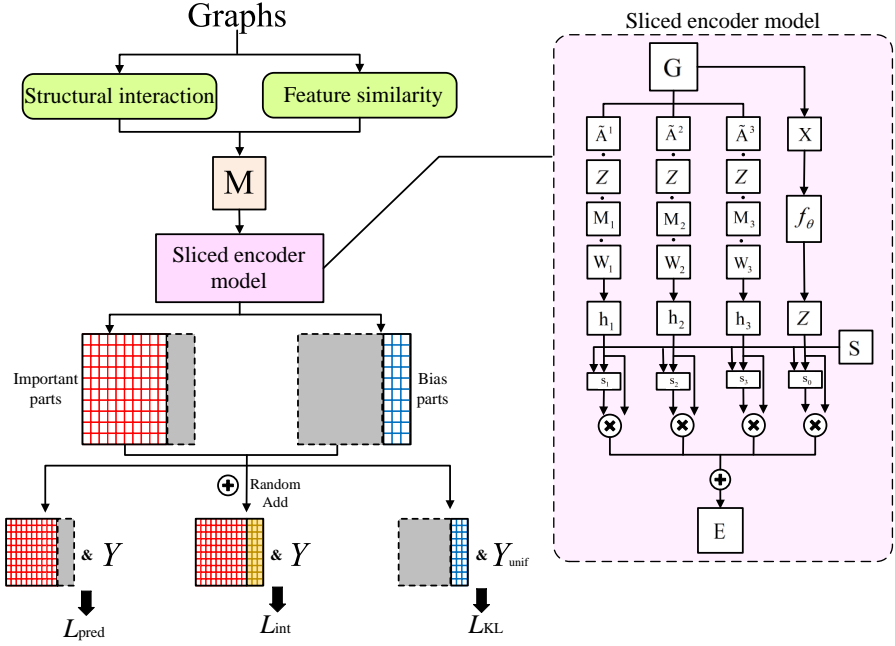}
    \caption{The model framework of CDM-GNN.}
    \label{Figure2}
\end{figure}

\subsection{Fine-grained closeness mask}
Based on the multi-granularity characteristics of graph data, this paper first considers modeling the nodes at a fine-grained level to capture the closeness between nodes, forming a mask matrix. 

Given a graph $G$, the attention values are calculated based on the feature similarity between node $i$ and node $j$ from a fine-grained perspective.
\begin{equation}
e_{ij}=\alpha ^{T}  \left ( W_{feat}\cdot x_{i} \parallel   W_{feat} \cdot x_{j}  \right )  \label{1}
\end{equation}
where $W_{feat}$ is the learnable parameters.
Then the calculated attention values are normalized.
\begin{equation}
\overline{e_{ij}}  =\frac{exp\left ( LeakyRelu\left ( e_{ij}  \right )  \right ) }{\sum_{m\in N_{i}}^{ }exp\left ( LeakyRelu\left ( e_{im}  \right )  \right )  }   \label{2}
\end{equation}

At the fine-grained level, we calculate the interaction between the structures of node $i$ and $j$ and use $Stru$ as the attention value of the topological structure to describe the relationship between nodes.
\begin{equation}
Stru_{ij} =\frac{\sum_{p\in N_{i}\cup N_{j}  }^{} min\left ( \omega _{ip},\omega _{jp}   \right ) }{\sum_{p\in N_{i}\cup N_{j}  }^{} max\left ( \omega _{ip},\omega _{jp}   \right )}  \label{3}
\end{equation}

Here, We use restart random walks to describe the degree of structural similarity between the center node $i$ and other nodes $p$. Particles start from the center node $i$ and randomly walk to their neighbors $p$ and $p\in N_{i}$. At each step, there is a certain probability of returning to the center node $i$. After $t$ iterations, the probability vector of visiting the neighbors around node $i$ is obtained.
\begin{equation}
\omega _{ip}^{t+1}=q\cdot \tilde{A}\omega _{ip}^{\left ( t \right ) } +\left ( 1-q \right )\cdot vec_{i}   \label{4}
\end{equation}
where $\tilde{A} =D^{-0.5} AD^{-0.5} $ and $D$ is the degree matrix. $q$ is the probability of restarting the random walk. $vec_i$ is a one-hot vector where the center node $i$ is assigned 1. Others are assigned 0. 

The probability vector is proportional to the edge weights. The higher the probability, the larger the edge weight. This probability vector is used as the weight vector. When $t\to \infty $, the vector converges to the following equation:
\begin{equation}
\omega _{ip}=\left ( 1-q \right )\cdot   \left ( I-\tilde{A} \right )^{-1}  \cdot e_{i}
\label{5}
\end{equation}

In Equation (3), $\omega_{ip}$ represents the weight of node $p$ ($p\in N{i}$). After normalizing $Stru_{ij}$, we obtain the following expression:
\begin{equation}
\overline{Stru_{ij} } = \frac{exp\left ( Stru_{ij}  \right ) }{\sum_{m\in N_{i} }^{} exp\left ( Stru_{im}  \right ) }   \label{6}
\end{equation}

We integrate $e_{ij}$ and $Stru_{ij}$ to obtain $M_{ij}$, resulting in the formation of matrix $M$:
\begin{equation}
M_{ij}=\frac{\overline{e_{ij}}+ \overline{Stru_{ij}}}{2}    \label{7}
\end{equation}

At a fine-grained level, this paper describes the closeness between each node from its features and structure. And it fuses them to form a mask. The mask part is an important substructure, while others are the bias parts.
\subsection{Coarse-grained disentangled framework}

Firstly, at a fine-grained level, the features are subjected to a simple feature transformation using the $f_{\theta }$ function with the learnable parameters $\theta$.
\begin{equation}
Z=f_{\theta } \left ( X \right ) \label{8}
\end{equation}
where the $f_{\theta }$ is a multi-layer neural network with $\theta$.

Next, the transformed features, fine-grained closeness mask, and adjacency matrices of different orders are sent into slice layers. The modeling process starts with the important substructures.

In the important substructures, the $M$ matrix is constructed in each slice GNN layer following the approach described in the previous section. In the first layer, $M_1$ and $\tilde{A} $ are used, in the second layer, the $M_2$ and $\tilde{A} ^{2} $ are used, and in the third layer, the $M_3$ and $\tilde{A} ^{3} $ are used. These matrices are then inputted into their corresponding slice layers to obtain the hidden layer states $h_1$, $h_2$, and $h_3$, and $W_{1}$,$W_{2}$, and $W_{3}$ are learnable parameters.
\begin{equation}
h_{n} = \tilde{A}^{n}  \cdot Z\cdot M_{n} \cdot W_{n},   \ \ \ \ n=1,2,3   \label{9} 
\end{equation}

The CDM-GNN model stacks the transformed feature results and obtained hidden layer states:
\begin{equation}
H=stack\left ( Z,h_{1},h_{2},h_{3}  \right )  \label{10}
\end{equation}

This model adaptive learns the weight $S$ for each slice layer:
\begin{equation}
S =reshape\left ( \sigma \left ( Hs  \right ) \right )  \label{11}
\end{equation}
\begin{equation}
E= squeeze\left ( SH \right )    \label{12}
\end{equation}

For the graph $G_{k} $, this model performs pooling on the obtained embedding to obtain the representation of $Emb_{k}$:
\begin{equation}
E_{k}= pooling\left(E \right)    \label{13}
\end{equation}

For bias parts, the CDM-GNN obtains the status of hidden layer $\overline{h_{1} }$, $\overline{h_{2}}$, and $\overline{h_{3}}$. In this model, we share these learnable parameters $W_{1}$,$W_{2}$, and $W_{3}$.
\begin{equation}
\overline{h_{n}}  = \tilde{A}^{n}  \cdot Z\cdot\left ( 1- M_{n}\right )   \cdot W_{n},   \ \ \ \ n=1,2,3    \label{14}
\end{equation}

Similarly, this model stacks the transformed feature results and obtained hidden layer states:
\begin{equation}
\overline{H} =stack\left ( Z,\overline{h_{1}} ,\overline{h_{2}} ,\overline{h_{3}}   \right )  \label{15}
\end{equation}

Then, the CDM-GNN model adaptive learns the weight $\overline{S } $ of each slice layer and gets the final embedding $\overline{Emb}_{k}$:
\begin{equation}
\overline{S}  =reshape\left ( \sigma \left ( H\overline{s}   \right ) \right )  \label{16}
\end{equation}
\begin{equation}
\overline{E} = squeeze\left ( \overline{S} H \right )   \label{17}
\end{equation}
\begin{equation}
\overline{E}_{k} = pooling\left ( \overline{E} \right )   \label{18}
\end{equation}

\subsection{Causal distangled learning}
This paper aims to train $K$ graphs by the causal components of their representations to enable the CDM-GNN model to classify correctly. To achieve this, CDM-GNN employs the supervised classification cross-entropy loss as follows:
\begin{equation}
    Y_{k}^{'}=softmax\left ( E_{k}  \right )   \label{19}
\end{equation}
\begin{equation}
L_{pred}=-  \sum_{k}^{}Y_{k}^{T} log\left ( Y_{k}^{'}   \right )  \label{20}
\end{equation}

For the representation of bias parts, it should not affect the classification results. Therefore, the prediction results of the bias parts should be evenly distributed across all categories. Using uniform distribution to help learn the representation of bias parts as follows: 
\begin{equation}
L_{KL}=  \sum_{k }^{}KL\left ( Y_{unif}  ,\overline{E}_{k}  \right ) \label{21}
\end{equation}
where $KL$ means the KL-Divergence, $Y_{unif}$ denotes the uniform distribution.

In order to reduce the interference of bias parts, the representation of bias parts is recombined with the causal parts to construct intervention terms. The CDM-GNN redefines the $\oplus $ function. It adds the bias representation of random disturbance back to the corresponding positions of the causal part representation to obtain the constructed intervention term $E_ {k} ^ {int}$:
\begin{equation}
E_{k}^{int}= E_{k} \oplus  \overline{E}_{k}   \label{22}
\end{equation}

In this case, the CDM-GNN can get the correct classification results by the representation of the causal part. The loss function is as follows:
\begin{equation}
    Y_{k}^{'int}=softmax\left ( E_{k}^{int}   \right )  \label{23}
\end{equation}
\begin{equation}
L_{int}=-\sum_{k }^{} Y_{k}^{T} log\left (Y_{k}^{'int}  \right )   \label{24}
\end{equation}

The overall loss function is as follows:
\begin{equation}
Loss= \frac{1}{\left | K \right | }\left ( \alpha L_{pred}+ \beta  L_{KL}+\gamma L_{int} \right )  \label{25}
\end{equation}
where $\alpha$, $\beta$, and $\gamma$ are hyper-parameters that determine the strength of disentanglement and causal influences. 
\section{Experiment}

\subsection{Datasets}
The commonly-used datasets for graph classification are summarized in Table 1.
\begin{itemize}
\item[$\bullet$]MUTAG \cite{debnath1991structure}: This dataset contains 188 compounds marked according to whether it has a mutagenic effect on a bacterium.
\item[$\bullet$]PTC \cite{DBLP:journals/bioinformatics/ToivonenSKKH03}: This dataset contains 344 organic molecules marked according to their carcinogenicity on male mice.
\item[$\bullet$]IMDB-M \cite{DBLP:conf/aaai/RossiA15}: This dataset is a movie collaboration dataset marked according to the genre an ego-network belongs to (romance, action, and science).
\end{itemize}
\begin{table}[H] 
\centering
\caption{Statistics of datasets}
\setlength{\tabcolsep}{5.5mm}\renewcommand{\arraystretch}{1.0}{
\begin{tabular}{|c|c|c|c|c|}
\hline
Dataset & Graphs & Classes & Avg.$|V|$ & Avg.$|E|$ \\
\hline
MUTAG & 188   & 2     & 17.93  & 19.79  \\
PTC   & 344   & 2     & 14.29  & 14.69  \\
IMDB-M & 1500  & 3     & 13.00  & 65.94  \\
\hline
\end{tabular}}
\label{table1}%
\end{table}%
\subsection{Baselines}
This paper compares the proposed CDM-GNN model with several state-of-the-art methods, which are summarized as follows:
\begin{itemize}
\item[$\bullet$] GCN \cite{DBLP:conf/iclr/KipfW17}: It is a semi-supervised graph convolution network model for graph embedding.
\item[$\bullet$] GAT \cite{DBLP:conf/iclr/VelickovicCCRLB18}: It is a graph neural network model which employs an attention mechanism to obtain graph representations.
\item[$\bullet$]Top-K \cite{DBLP:conf/icml/GaoJ19}: It is a graph representation method that adaptively selects some critical nodes to form smaller subgraphs based on their importance vectors.
\item[$\bullet$]ASAPool \cite{DBLP:conf/aaai/RanjanST20}: It is a graph neural network model that utilizes attention mechanisms to capture the importance of nodes and pools subgraphs into a coarse graph through learnable sparse soft clustering allocation for graph representation.
\item[$\bullet$]SUGAR \cite{DBLP:conf/www/SunLPWNY021}: It is a graph representation method that first samples some subgraphs. And then it uses the DQN algorithm to select top-k key subgraphs as representative abstractions of the entire graph.
\item[$\bullet$]SAT \cite{DBLP:conf/icml/ChenOB22}: It is a graph transformer method that incorporates the structure explicitly. Before calculating attention, it fuses structural information into the original self-attention by extracting k-hop subgraphs or k-subtrees on each node.
\end{itemize}
\subsection{Performance on Real-world Graphs}
In the graph classification task, this paper adopts accuracy as the evaluation metric to measure the performance of different models.

From Table 2, it can be observed that the proposed CDM-GNN model achieves the best graph classification results on the MUTAG, PTC, and IMDB-M datasets. Specifically, compared with the GCN and GAT, the CDM-GNN has an improvement of about 5$\%$ on the MUTAG dataset in terms of graph classification accuracy. Compared to Top-K and ASAPool, which are specifically designed for pooling operations, CDM-GNN demonstrates improvements ranging from 12$\%$ to 20$\%$. In comparison to methods focusing on substructure extraction, SUGAR and SAT, CDM-GNN shows improvements of 3$\%$ to 7$\%$. On the PTC dataset, the CDM-GNN also shows a notable improvement and achieves accurate graph classification. On the multi-class IMDB-M dataset, the proposed CDM-GNN achieves about 5.4$\%$ improvement over GCN and GAT. When compared to some pooling methods, CDM-GNN shows accuracy improvements ranging from 5.5$\%$ to 15$\%$. In the substructure research, CDM-GNN demonstrates accuracy improvements of 4$\%$ to 9$\%$.
These results highlight the superior graph classification performance of the CDM-GNN model across different datasets. It surpasses other popular models and specialized approaches for pooling or subgraph analysis.
\begin{table}[H]
  \centering
  \caption{The Accuracy of graph classification}
  \setlength{\tabcolsep}{8.0mm}
  \renewcommand{\arraystretch}{1.2}{
    \begin{tabular}{|c|c|c|c|}
    \hline
    Models & MUTAG & PTC   & IMDB-M \\
    \hline
    GCN   & 0.8924 & 0.5726 & 0.5700  \\
    GAT   & 0.8994 & 0.5944 & 0.5810  \\
    Top-K Pool & 0.7291 & 0.5721 & 0.4836 \\
    ASAPool & 0.8211 & 0.5677 & 0.5794 \\
    SUGAR & 0.8660  & 0.5821 & 0.5988 \\
    SAT   & 0.9030  & 0.6070  & 0.5329 \\
    Ours  & \textbf{0.9474} & \textbf{0.6154} & \textbf{0.6348} \\
    \hline
    \end{tabular}}
  \label{table2}%
\end{table}
\subsection{Visualization results}
To demonstrate the interpretability of the CDM-GNN method, this paper conducts a qualitative analysis. According to the existing chemical knowledge, people know that the MUTAG dataset determines whether it has a mutagenic effect by judging whether the molecule contains the substructure of $NO_{2} $ or $NH_{2}$. Therefore, this paper visualizes the MUTAG dataset to observe the interpretability of the CDM-GNN model and perform qualitative analysis. If CDM-GNN model can recognize the important substructure $NO_{2}$ and disentangle this part with bias parts, then it indicates that the model has better interpretability. This model can obtain classification and interpretation results that are consistent with human cognition and prior knowledge. 

\begin{figure}
\centering
\subfigure[Original 123]{\includegraphics[width=0.3\textwidth]{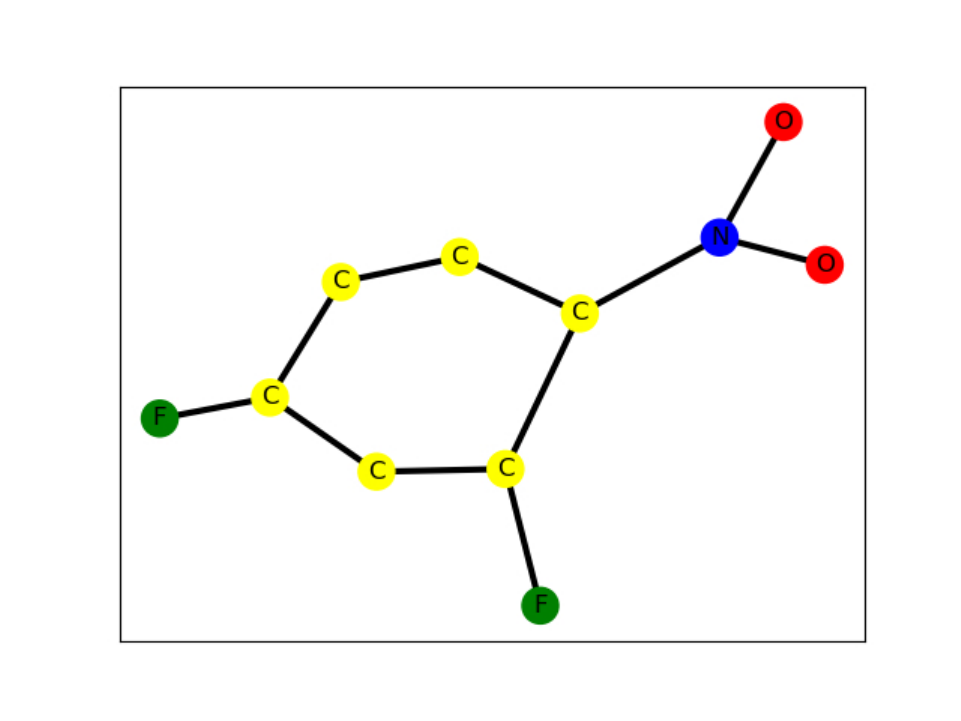}} 
\subfigure[Original 31]{\includegraphics[width=0.3\textwidth]{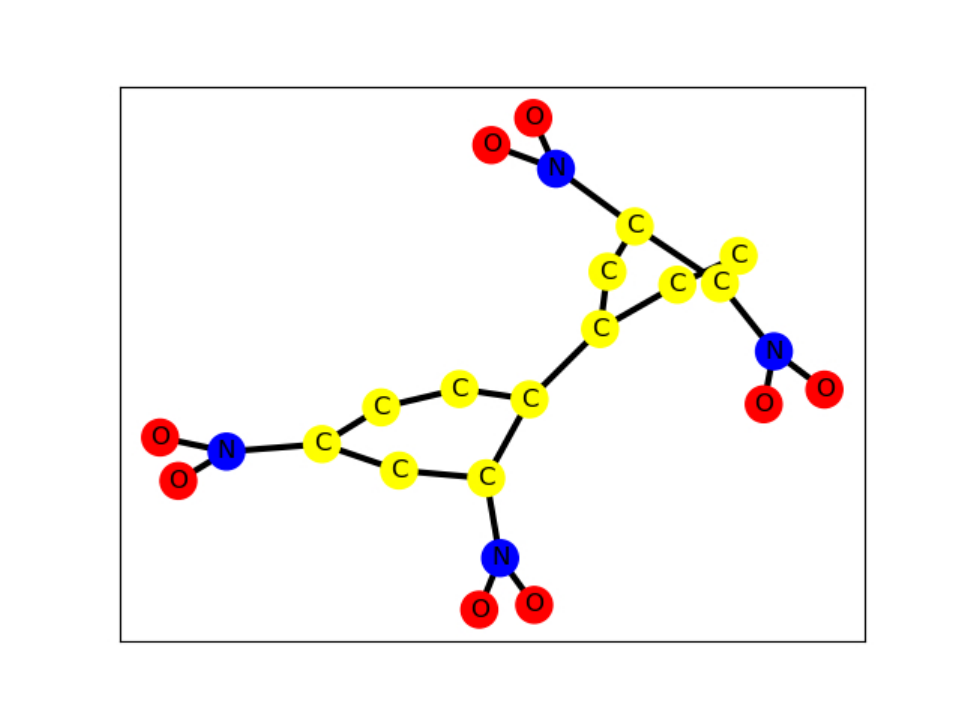}}
\subfigure[Original 164]{\includegraphics[width=0.3\textwidth]{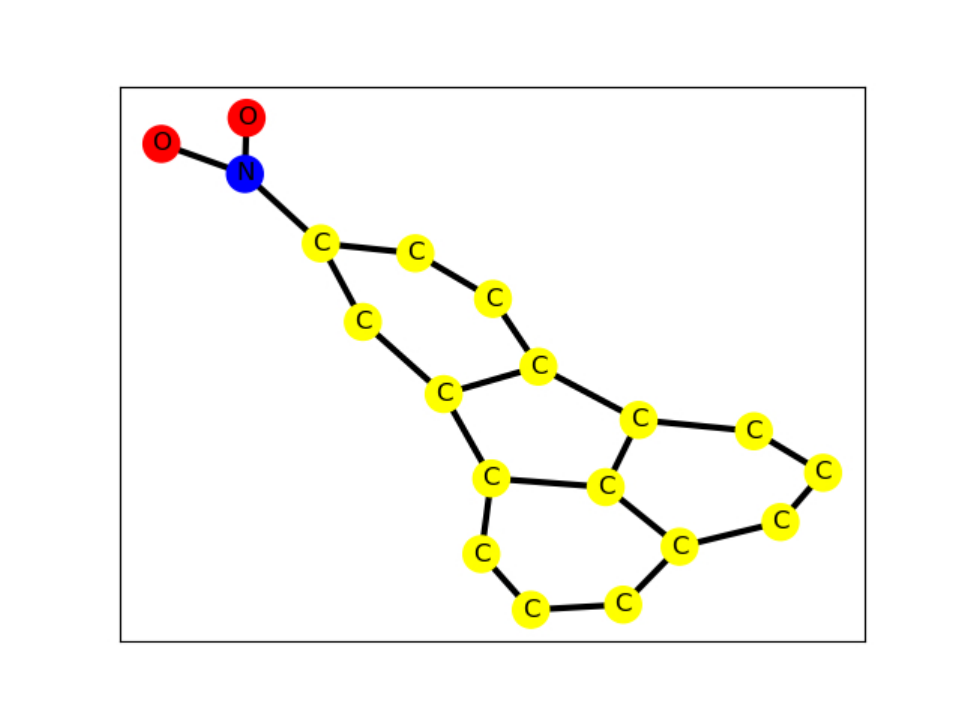}}
\\ 
\centering
\subfigure[Disentangle 123]{\includegraphics[width=0.3\textwidth]{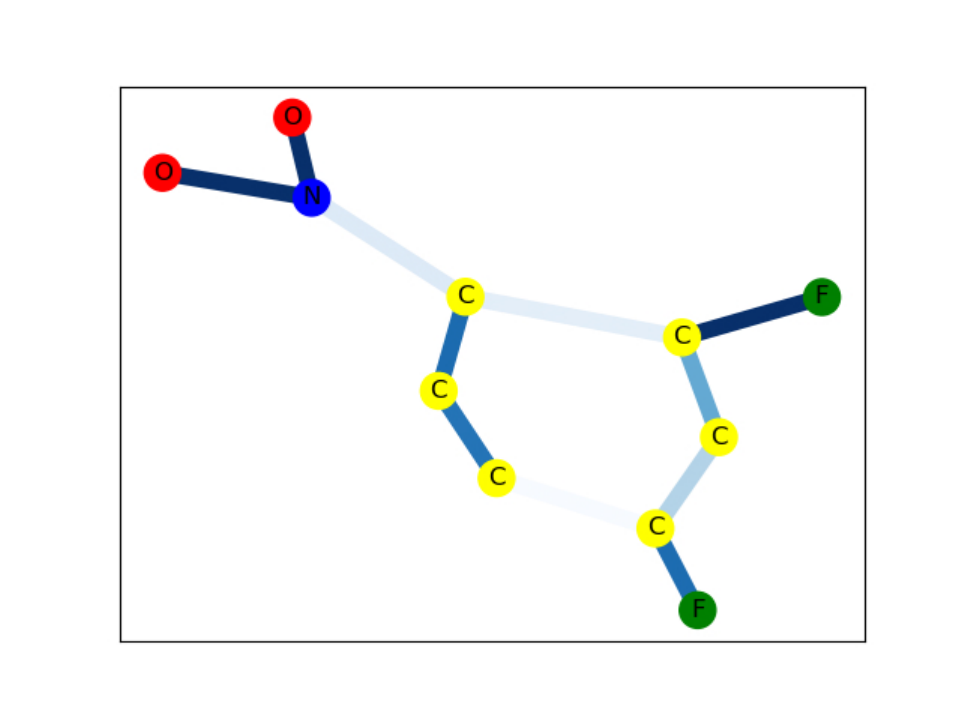}}
\subfigure[Disentangle 31]{\includegraphics[width=0.3\textwidth]{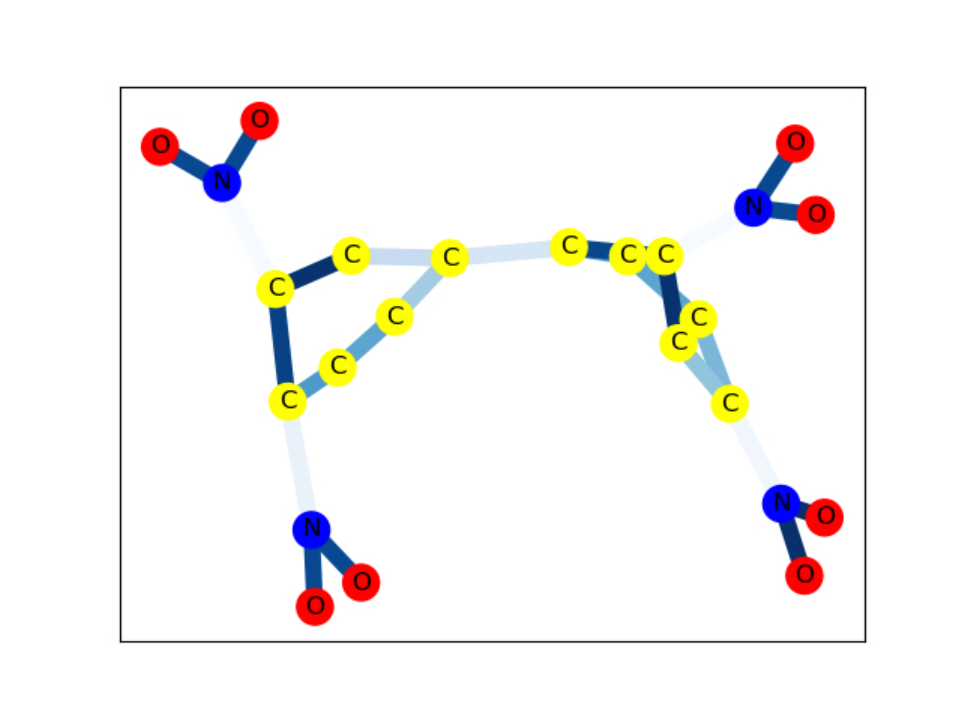}}
\subfigure[Disentangle 164]{\includegraphics[width=0.3\textwidth]{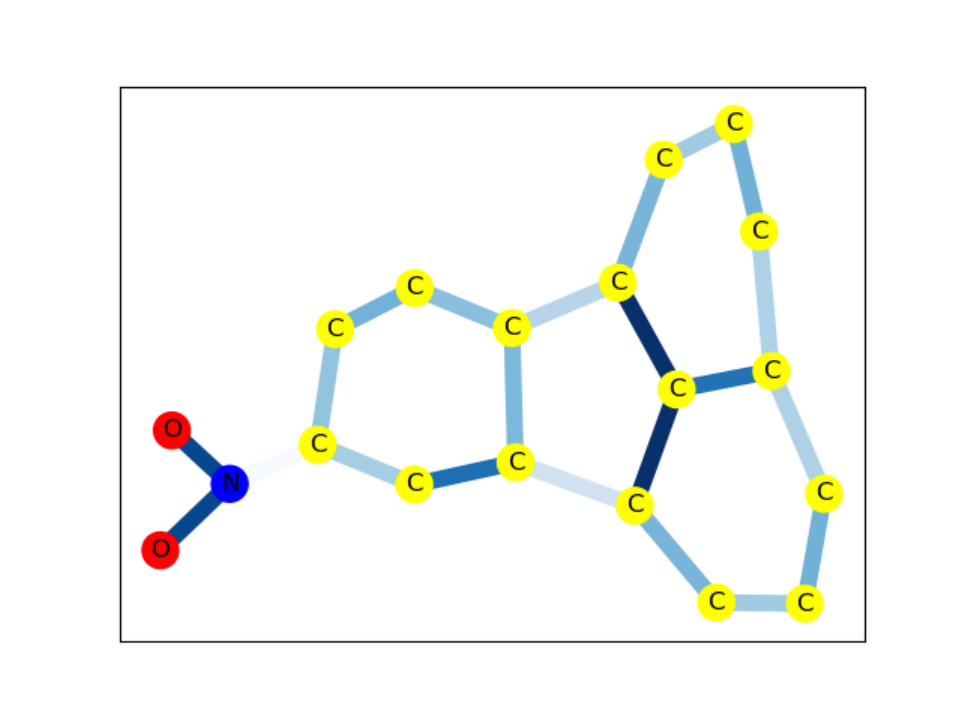}}
\caption{The visualization results on the MUTAG dataset.} 
\label{figure3}
\end{figure}

In this study, the edges of the graph are colored based on the values of the mask, where darker colors indicate greater importance. In the visualized results, we can observe that the edges of important substructures $NO_{2}$ are colored darker, indicating their higher weights and greater significance in the classification of the MUTAG dataset. The edges connecting the important substructures $NO_{2}$ and the bias parts are colored lighter, indicating that the CDM-GNN model has successfully disentangled the important substructures from the bias parts. From Figure 3 (a)(b)(d)(e), it can be observed that regardless of whether the graph contains one or multiple $NO_{2}$ substructures, the CDM-GNN can recognize them and successfully disentangle them from the carbon rings. Even in cases where there are multiple carbon rings in the graph, like Figure 3 (c)(f), the CDM-GNN model can still identify the important $NO_{2}$ parts. This indicates the model has the ability to capture and distinguish the relevant features, enabling it to recognize and disentangle the important $NO_{2}$ parts from other structural components.

\subsection{Ablation experiment}
Here, this paper discusses the role of three losses in learning for the CDM-GNN model.
From Figure 4, It can be observed that when this model only uses the $L_{KL}$, $L_{int}$, and $L_{pred}$, the $L_{pred}$ plays a more important role in this model training. The CDM-GNN model is designed by incorporating downstream graph classification tasks. The $L_{pred}$ plays a leading role in guiding this model to achieve better differentiation among different categories of graph data. This way, the CDM-GNN model focuses on the goal of graph classification and progressively improves its performance. The $L_{KL}$ is mainly aimed at reducing the interference caused by the bias component. Without this loss, the classification results are the most significant decrease in accuracy. The $L_{int}$ is to consider the decoupling of causality. If the CDM-GNN model is without $L_{int}$, it also leads to a decrease in accuracy. When three losses are used together, the model can achieve the best effect. It has better graph classification results and can separate the important substructure and the bias parts.
\begin{figure}[h]
\centering
\includegraphics[width=0.7\textwidth]{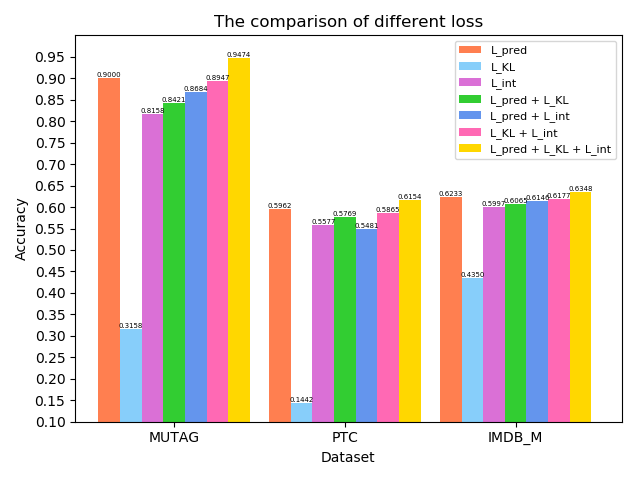}
\caption{The ablation results about the Loss on MUTAG, PTC, and IMDB-M dataset} 
\label{figure4}
\end{figure}
\section{Conclusion}
This paper introduces a causal disentangled multi-granularity graph representation learning method, namely CDM-GNN. It is primarily based on the idea of multi-granularity, aiming to identify important substructures and bias parts and disentangle them in the graph classification tasks. The proposed method achieves favorable classification performance and provides qualitative interpretability of the results. 
This technology can be extended for researching drug molecule properties and facilitating new drug development. Identifying important substructures that influence drug properties within molecules and their disentanglement can aid in exploring drug molecule properties and innovating drug development based on these substructures.
This paper has not yet explored additional downstream graph tasks, such as node classification and link prediction. Future research could explore the extension of the concept of causal disentangled multi-granularity to these graph-related tasks. Subsequent research also can prioritize in-depth discussions of interpretable quantitative assessments for self-explainable models.

\subsubsection{Acknowledgments} This work is supported by the National Natural Science Foundation of China (Nos. 62221005, 61936001, 61806031), Natural Science Foundation of Chongqing, China (Nos. cstc2019jcyj-cxttX0002, cstc2021ycjh-bgzxm0013), Project of Chongqing Municipal Education Commission, China (No. HZ2021008), and Doctoral Innovation Talent Program of Chongqing University of Posts and Telecommunications, China (Nos. BYJS202108, BYJS202209, BYJS202118).

%
%

%
%
%
%
%
\end{document}